# Transformer-based Conditional Generative Adversarial Network for Multivariate Time series Generation

**Abdellah Madane**
madane@lipn.univ-paris13.fr

**Mohamed-Djallel DILMI**
dilmi@lipn.univ-paris13.fr

**Florent Forest**
forest@lipn.univ-paris13.fr

**Hanane AZZAG**
azzag@lipn.univ-paris13.fr

**Mustapha Lebbah**
Mustapha.lebbah@uvsq.fr

**Jérôme Lacaille**
jerome.lacaille@safrangroup.com

## Abstract

Conditional generation of time-dependent data is a task that has much interest, whether for data augmentation, scenario simulation, completing missing data, or other purposes. Recent works proposed a Transformer-based Time series generative adversarial network (TTS-GAN) to address the limitations of recurrent neural networks. However, this model assumes a unimodal distribution and tries to generate samples around the expectation of the real data distribution. One of its limitations is that it may generate a random multivariate time series; it may fail to generate samples in the presence of multiple sub-components within an overall distribution. One could train models to fit each sub-component separately to overcome this limitation. Our work extends the TTS-GAN by conditioning its generated output on a particular encoded context allowing the use of one model to fit a mixture distribution with multiple sub-components. Technically, it is a conditional generative adversarial network that models realistic multivariate time series under different types of conditions, such as categorical variables or multivariate time series. We evaluate our model on UniMiB Dataset, which contains acceleration data following the XYZ axes of human activities collected using Smartphones. We use qualitative evaluations and quantitative metrics such as Principal Component Analysis (PCA), and we introduce a modified version of the Frechet inception distance (FID) to measure the performance of our model and the statistical similarities between the generated and the real data distributions. We show that this transformer-based CGAN can generate realistic high-dimensional and long data sequences under different kinds of conditions.

## 1 Introduction

Conditional generative adversarial networks have attracted significant interest recently (Hu et al., 2021; Liu & Yin, 2021; Liu et al., 2021). The quality of generated samples by such models is improving rapidly. One of their most exciting applications is multivariate time series generation, particularly when considering contextual knowledge to carry out this generation. Most published works address this challenge by using recurrent architectures (Lu et al., 2022), which usually struggle with long time series due to vanishing or exploding gradients. One other way to process sequential data is via Transformer-based architectures. In the span of five years, Transformers have repeatedly advanced the state-of-the-art on many sequence modeling tasks (Yang et al., 2019)(Radford et al., 2019)(Conneau & Lample, 2019). Thus, it was a matter of time before we could see transformer-based solutions for time series (Yoon et al., 2019)(Wu et al., 2020)(Mohammadi Farsani & Pazouki, 2020), and particularly for multivariate time series generation (Li et al., 2022)(Leznik et al., 2021). These studies showed promising results. Consequently, whether Transformer-based techniques are suitable for conditional multivariate time series generation is an interesting problem to investigate. Our work extends the TTS-GAN (Li et al., 2022) by conditioning its generated output on a particular encoded context allowing the use of one model to fit a mixture distribution with multiple sub-components. Our contributions are summarized as follows:

- We designed a transformer-based conditional generative adversarial network architecture for conditional multivariate time series generation using different type of conditions.
- We introduce and study a new parameter, alpha $\alpha$, which controls the percentage of the desired noise/variability and the percentage of relevance given to the context in a conditional generative adversarial network.
- We rigorously evaluate our approach on the UniMiB SHAR dataset using different qualitative and quantitative evaluation methods.
- We introduce MTS-FID, a version of FID suitable for evaluating generative models dealing with multivariate time series. We show that MTS-FID exhibits the same behavior as FID and that the results obtained correlate with other evaluations performed.
- Using a data augmentation study, we demonstrate that our Transformer-based CGAN could accurately model the mixture distribution of the classes given as a condition and generate precise multivariate time series for each class.

## 2 RELATED WORKS

**Conditional Generative Adversarial Networks** (CGANs) (Mirza & Osindero, 2014) are an extension of vanilla GANs (Goodfellow et al., 2014). They allow to obtain a certain degree of control over the generated samples. A CGAN model initially sets a condition that the generated data must meet. This condition can take different forms: it can be the class of the image we want to generate in the case of a GAN model trained for an image generation task, or some context encoding for a time series generative model. CGAN extends the vanilla GAN architecture by incorporating a condition in the form of a vector $c$ concatenated with the noise vector $z$ at the input of the generator and provided as an input to the discriminator. Thus, this corresponds to conditioning the $G(z)$ and $D(x)$ distributions. Therefore, the standard loss function of CGAN is as follows:

$$L(D, G) = \mathbb{E}_{x \sim p_{\text{data}}} [\log D(x \mid y)] + \mathbb{E}_{z \sim p_z} \left[ \log \left( 1 - D(G(z \mid y)) \right) \right] \quad (1)$$

**Transformers**, as presented in the seminal paper Vaswani et al. (2017), is an architecture relying entirely on the attention mechanism to learn global dependencies between input and output, without any notion of recurrence. Following this work, we realized that we could build architectures using only attention mechanisms, which proved to be sufficient to understand and extract features from the input given to the model. The subtlety here, and what makes a difference compared to a recurrent network (RNN), is that Transformers process every sequence elements simultaneously, in parallel. Also, the attention mechanism ensures the possible use of any element of the sequence whenever we compute attention for it. Thus, it does not lose any relevant information whatsoever.

**Transformer-based Generative Adversarial Networks**

**TransGAN** (Jiang et al., 2021) introduces, for the first time, a generative adversarial network built using solely Transformers that generates synthetic images. Like other GANs, the TransGAN consists of two parts: a generator and a discriminator; first, the generator takes a one-dimensional noise vector as an input and gradually increases the resolution of the feature map computed using transformer encoder blocks until a synthetic image with the required resolution is generated. As for the discriminator, the authors adopted the exact model for Vision Transformer (ViT) (Dosovitskiy et al., 2020a), an image classifier based on Transformers.

**TTS-GAN** (Li et al., 2022) is a transformer-based GAN where the generator and the discriminator adopt the Trans-encoder architecture. It generates synthetic multidimensional time series for data augmentation purposes. The process followed was similar to the TransGAN for image generation. Authors of TTS-GAN consider a multivariate time series as an image of height equal to one and length equal to the number of time steps and the number of variables in the multivariate time series as the number of channels.

## 3 MTS-CGAN: MULTIVARIATE TIME SERIES CONDITIONAL GENERATIVE ADVERSARIAL NETWORK

We propose a conditional generative adversarial network where the generator and the discriminator are purely transformer-based neural networks. This model generates multivariate time series given a context. This context could be categorical variables or a multivariate time series. The model is discussed and evaluated in detail in the following sections.

## 3.1 ARCHITECTURE

MTS-CGAN architecture is inspired by the models introduced in Jiang et al. (2021); Dosovitskiy et al. (2020b); Li et al. (2022). Like other GANs, it consists in a generator (*G*) and a Discriminator (*D*), as shown in Figure 1.

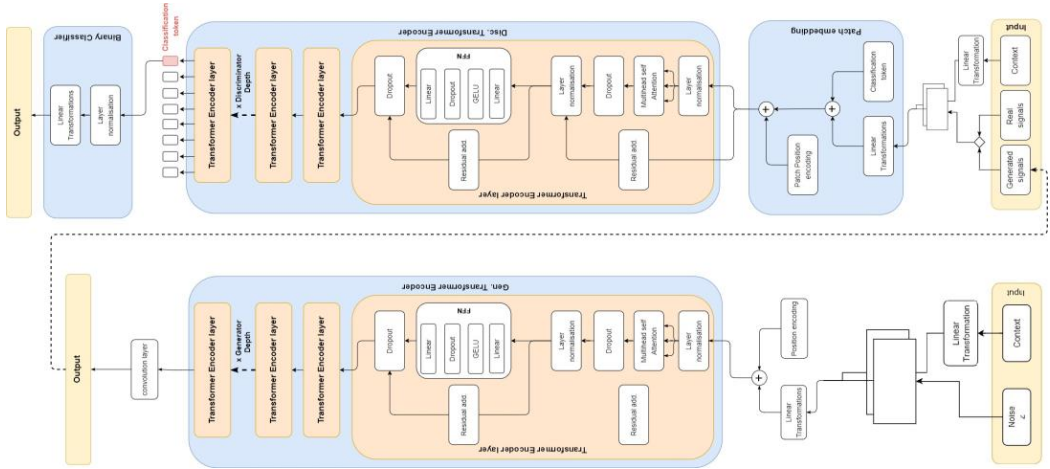

Figure 1: MTS-CGAN architecture.

**The conditional generator** has two inputs: the random noise vector of dimension $d_z$ and the encoded context to condition the generation. The latent dimension is a hyper-parameter depending on the dataset. Also, the context is encoded to fit a latent space of dim $d_z$ to facilitate its concatenation with the noise vector *z*. We then apply linear transformations to the concatenated vectors to obtain a vector of size equal to the target sequence length and with *c* channels, where *c* needs to be tuned. Finally, we encode the position of the elements of this vector. The resulting vector passes through the consecutive layers of a Transformer encoder. Each has a multi-head self-attention layer that extracts the contextual inter-dependencies between the generated signal and the provided context. The final output of those layers passes through a (1, 1)-convolution layer with a filter size equal to the dimension of the time series we are aiming to generate.

**Alpha $\alpha$** : Every modeling deals with the object of interest – in our case, a multivariate time series – as a composition of two components: a regular and an irregular one. Contextual knowledge and prior information encode evidence in favor of some propositions and help to characterize the regular component of the time series while a random variable models the irregular one. Thus, it seems natural to associate a weight $\alpha$ with the random variable that models the lack of knowledge. This hyper-parameter $0 < \alpha < 1$ defines the percentage of the desired noise/variability and the percentage of relevance given to the context by complementary. Therefore, we multiply the noise *z* by $\alpha$ and the context vector by $1 - \alpha$.

**The conditional discriminator** aims at classifying the time series as real or synthetic and takes as input either a real time series with its corresponding context, or a generated one. First, it concatenates the input vectors and applies a linear transformation. Then, the resulting embedding is broken down into multiple patches associated with their respective positions and a classification token. The whole set is then fed to the consecutive layers of the Transformer's encoder, as was the case in the generator. Finally, a binary classifier uses the information embedded into the classification token to distinguish between the real and fake signals.

## 3.2 Loss function

We trained MTS-CGAN using different loss functions: standard CGAN loss (Goodfellow et al., 2014) (see Eq. 1), Least Squares GAN (LSGAN) (Mao et al., 2017), and Wasserstein GAN with Gradient Penalty (WGAN-GP) (Gulrajani et al., 2017).

**LSGAN** replaces the sigmoid cross-entropy loss used in the original GAN with the least squares loss to train both the generator and the discriminator. The idea behind this proposal is to be able to penalize the generated samples that are far away from the decision boundary of the discriminator and bring them closer to the actual distribution of the data. Therefore, the generator will learn to match the real distribution and the only time the gradient of the discriminator will be equal to zero is when the generator captures the data manifold.

$$\min_D V_{\text{LSGAN}}(D) = \frac{1}{2}\mathbb{E}_{x \sim p_{\text{data}}(x)}[(D(x \mid y) - b)^2] + \frac{1}{2}\mathbb{E}_{z \sim p_z(z)}[(D(G(z \mid y)) - a)^2] \quad (2)$$

$$\min_G V_{\text{LSGAN}}(G) = \frac{1}{2}\mathbb{E}_{z \sim p_z(z)}[(D(G(z \mid y)) - c)^2] \quad (3)$$

Where $a$ represents labels for fake data, $b$ represents labels for real data, $c$ refers to the value that the generator (G) wants the discriminator (D) to believe for fake data, and $y$ is the condition.

**WGAN-GP** (Gulrajani et al., 2017) is an improved version of the WGAN Arjovsky et al. (2017). The weight clipping method introduced in the WGAN to enforce the Lipschitz constraint is not the best solution. An alternative is proposed. It consists in computing a gradient penalty (GP), which is added to the "loss" of the discriminator to maintain a norm of its gradients close to 1.

$$\mathcal{L}_D^{\text{WGAN-GP}} = \mathcal{L}_D^{\text{WGAN}} + \lambda \mathbb{E}_{z \sim p_z}\left[\left(\left\|\nabla_z D(G(z \mid y))\right\|_2 - 1\right)^2\right] \quad (4)$$

$$\mathcal{L}_G^{\text{WGAN-GP}} = -\mathbb{E}_{z \sim p_z}[D(G(z))] \quad (5)$$

## 4 Experiments

### 4.1 Dataset

UniMiB SHAR (Micucci et al., 2017) is a human activity recognition and fall detection dataset. It consists of acceleration samples collected with Android smartphones for nine (9) types of activities of daily living (ADL) and eight (8) types of falls performed by 30 subjects ranging from 18 to 60 years old. In our study, we will focus on generating samples of activities of daily living. Since there is an imbalance in the number of samples between the different classes, we decided to keep only three ADL classes: Walking, Running and Going downstairs. These classes make up 4034 samples out of 11 771 samples of both human activities and fall in this dataset. Each of these classes has specific characteristics in its acceleration time series.

#### 4.1.1 Activity of daily living (ADL) as a categorical condition

Our first generation task considers activities of daily living class (ADL) as the condition. MTS-CGAN generates time series for each particular activity as shown on Figure (a) 2. We encode each class as a one-hot numerical array. We then apply some linear transformations and concatenate the output to the noise $z$.

#### 4.1.2 Multivariate time series as a condition

Our second generation task consists of generating time series of the z-axis based on the condition of multivariate time series of the x-axis and y-axis as shown on Figure (b) 2. In this case, we only used the data set of Running activity. We encode the time series of the (x, y) axes and use linear transformations to project them into the latent z-space dimension. Then we concatenate the new vector to the $z$-noise.

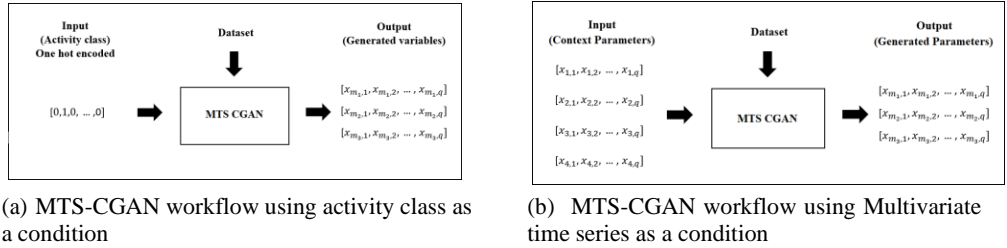

(a) MTS-CGAN workflow using activity class as a condition

(b) MTS-CGAN workflow using Multivariate time series as a condition

Figure 2: MTS-CGAN workflow

## 4.2 ALPHA RANGE

$\alpha$ represents the lack of knowledge. We trained MTS-CGAN using different values of $\alpha$ to evaluate this hyperparameter's relevance and impact. We tested values from the range (0.1, 0.25, 0.7, 0.9). We expect that low $\alpha$ values confer vast inter-classes differences between the different classes and low intra-class variability/inertia, while high $\alpha$ values tend to reduce inter-class differences and enhance intra-class variability/inertia.

## 5 RESULTS

### 5.1 ALPHA DISCUSSION

As mentioned, intra-class inertia represents the intra-class variability within classes. It may increase (resp. decrease) with high (resp. small) $\alpha$ values. Also, since inter-class inertia is correlated to prior knowledge, small alpha values may provide distant class kernels. Figure 3 shows the resulting visualization with t-SNE of real and 3-class conditioned generated Multivariate time series for the four MTS-CGAN models trained with different values of $\alpha$. the Global analysis of the four scatterplots corroborates our expectation. Indeed, for low values (0.1, 0.25) of $\alpha$, inter-class distances for the generated time series (Yellow, Green, and Blue) are much more important in comparison with those obtained for high values (0.75, 0.9). Also, since the hyperparameter $\alpha$ weights the random noise and the irregular component of time series, it's natural that low values (0.1, 0.25) of $\alpha$ give less scattered sets while high values (0.75, 0.9) manifest more scattered sets.

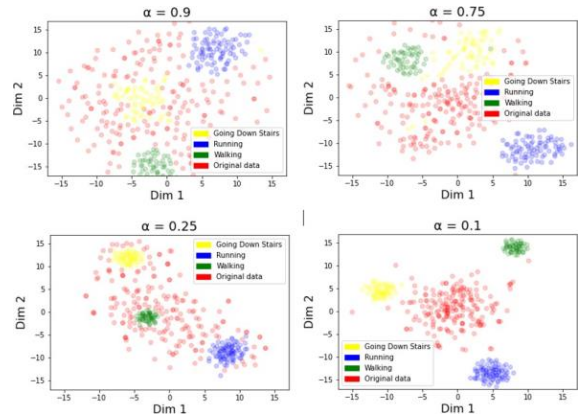

Figure 3: T-distributed stochastic neighbor embedding (TSNE) to represent real and fake time series of ADLs for the four $\alpha$ values. Red dotes represent samples from UniMiB dataset, (Blue,yellow and green dotes represent generated time series for the three classes (walking, GDS, running)

## 5.2 EVALUATION USING STATISTICAL METHODS

### 5.2.1 PRINCIPAL COMPONENT ANALYSIS

Based on a properly set-up and a solid PCA, we can project the generated data on the two-dimensional principal components (PC) space and compare it to the real data distribution. We can look at the figures and analyze precisely the proximities between points of each ADL.

Using figure 4, we visualize the projection of the generated data onto the factor space corresponding to the PCA of each axis (x,y,z). Recall that we have restricted the number of projected data depending on their quality of representation. For this reason, we have an unbalance between the data quantity in each visualized distribution. The first three figures (1-3) show that the distribution of the generated data (points in black) overlaps with the real data according to the three axes. It proves that the MTS-CGAN can generate synthetic data that looks very similar to the real data. The last three figures (4-6) allow us to evaluate the quality of the time series generated for each activity separately according to each axis. The x-axis shows that the data generated for the three activities perfectly matches the distribution of each activity class in the real data, that is, going downstairs and walking activities forming an ellipse in the middle and then the running class scattered all around. Therefore, the time series generated for the x-axis are similar to the real ones for each activity. Also, along the y-axis, generated data for going downstairs and walking activities are projected similarly and close to their real data distribution. The synthetic running activity data overlap with real running activity data. However, they do not cover the entire real distribution, which conveys a lack of diversity in the generation of this activity. The generator has learned to generate a particular type of running activity data. Also, let us not forget that some generated points are not visualized because of their poor representation quality, which may cover other areas of the real distribution of this activity. As for the z-axis, the distributions of the two activities going downstairs and walking match their corresponding distribution in real data, but the model lacks diversity for their generation. As with the Running activity, we cannot compare the projected data to the actual distribution along this axis due to their poor representation quality. We would rather rely on a decent and robust evaluation tool than an uncertain one.

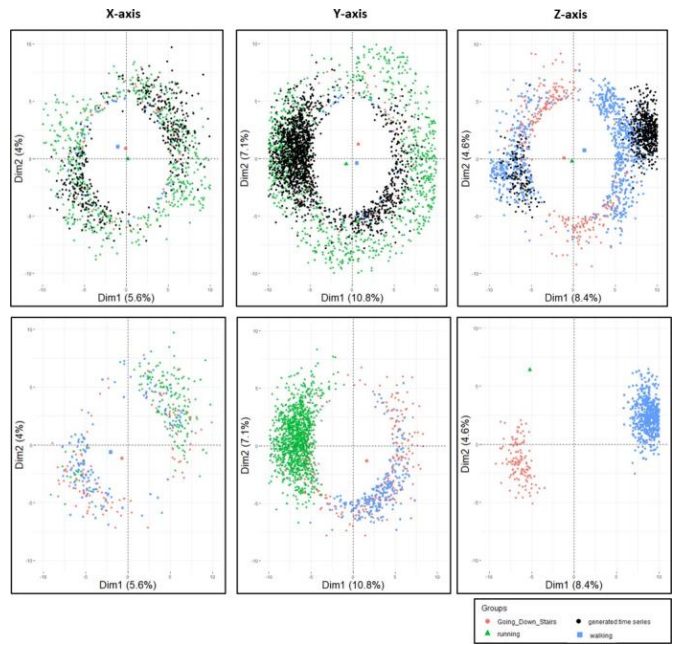

Figure 4: Projecting generated data by MTS-CGAN on real data PCA along each axis (*X, Y, Z*)

5.2.2 FREQUENCY DISTRIBUTION OF STATISTICAL FEATURES

Therefore, we solidify our evaluation using other statistical features extracted from the multivariate time series: the median, mean, and standard deviation. This method has already been used to evaluate other CGANs for multivariate time series generation (Chen et al., 2021). Since we normalized our training dataset, we assume that the generated time series will also have a mean equal to 0 and a standard deviation equal to 1 if the generator reproduces the real data distribution correctly. We are considering some slight variations as stated by the central limit theorem. The results show that the frequency distributions of the means and standard deviations of the synthetic time series produced by our generator show that they have a standard deviation of 1 and a mean of about 0. Also, it show that the frequency distribution of the median of real and generated data are similar along each activity and each axis.

5.3 FRECHET INCEPTION DISTANCE (FID)

The objective of a generative model is to produce outputs similar to the real data. Therefore, the distance between the generated and real data distribution can be used to measure the model's performance. A well-performing metric to evaluate the quality of GANs is the Frechet Inception Distance (FID) (Heusel et al., 2017). It has been used to evaluate many recent generative models and mainly to evaluate the quality of images generated by such models (Karras et al., 2019; 2020). It calculates the distance between the real distribution and the generated distribution based on statistics computed from a set of samples of each distribution. There are three steps to compute the FID score:

1. A pre-trained model, InceptionV3, is loaded without its classification layer for feature extraction purposes. It embeds the collection of real data samples and the collection of generated data samples into its high dimensional feature space.

2. Each collection of feature vectors is summarized as a multivariate Gaussian by calculating its mean and covariance.

3. The distance between the two gaussian is computed using Fréchet distance (Fréchet, 1957).

Considering ($\mu_r$, $\Sigma_r$) and ($\mu_g$, $\Sigma_g$) are respectively the mean and covariance of the sample embeddings from the real data distribution and the generated data distribution, the FID is formulated as follows :

$$FID = \|\mu_r - \mu_g\|^2 + T_r\left[\Sigma_r + \Sigma_g - 2(\Sigma_r\Sigma_g)^{1/2}\right] \quad (6)$$

The lower the FID score is, the better the generated output is.

5.3.1 FID FOR MULTIVARIATE TIME SERIES

Our task consists of multivariate time series generation. However, InceptionV3 is trained on ImageNet, which makes it suitable for image feature extraction. Therefore, FID, as used with InceptionV3, does not fit our case. We propose to replace InceptionV3 with a feature extraction model adapted to multivariate time series. Note that some previous works (Smith & Smith, 2020) have proceeded likewise to adapt the FID to their data type. As a result, we train a classifier on our dataset to classify multivariate time series (axes - XYZ) according to their corresponding activity (Walking, Running, Jumping, Going downstairs, Sitting down, Going upstairs, Standing up from sitting, Lying down from standing, Standing up from laying). It is a convolutional neural network model inspired by (Wang et al., 2017). We obtained better performances than the models proposed for the same classification task in the original paper of the dataset. We use this model to perform the feature extraction step in the FID by eliminating the classification layer and keeping the other ones. For the generation case where the condition consists of multivariate time series, we trained a z-axis series classifier instead of the three axes. We test our new FID version (MTS-FID) by computing the score between a collection of real samples and itself, which we expect to be 0. We then introduce a white gaussian noise to the collection of real samples and compute the MTS-FID between the two sets. We expect a higher score each time we increase the noise power. Figure 5 shows that, similar to the FID application for image generation, lower MTS-FID scores correlate with better-quality multivariate times series.

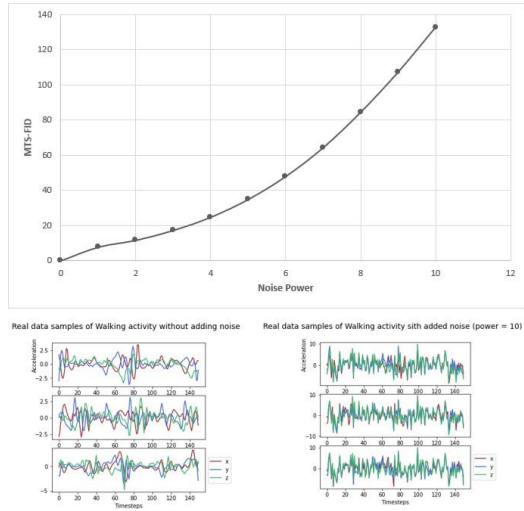

Figure 5: Behavior of MTS-FID when adding white gaussian noise

### 5.3.2 Evaluation using MTS-FID

We use MTS-FID to monitor the training of our models. It helps us know when they converge so we can stop the training and avoid overfitting. Figure 6 show the MTS-FID evolution during MTS-CGAN training. It starts with a high score since, in the beginning, the generator produces noise, then it gradually decreases as the generator becomes more capable of generating realistic data. The orange dotted line shows the exact epoch when the model reached the smallest value of MTS-FID during its training and thus its convergence. Using the previous evaluation methods, we find that stopping the training at this stage also gives the best results. This shows the correlation between the different evaluation methods. We also use MTS-FID to compare models trained using different hyperparameters, alpha values, and loss functions, as shown in table 1 and table 2.

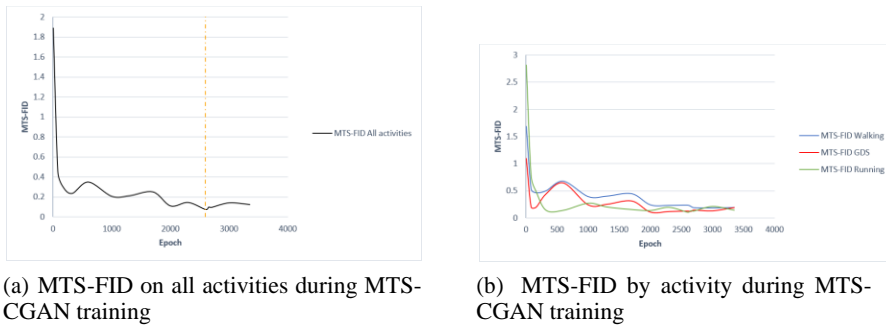

(a) MTS-FID on all activities during MTS-CGAN training

(b) MTS-FID by activity during MTS-CGAN training

Figure 6: MTS-FID during MTS-CGAN training

Experiments conducted on the first generation task and presented in table 1 were achieved with $\alpha = 0.9$, and experiments in table 2 were achieved using LSGAN loss function. Results show that MTS-CGAN generates better quality data when trained on LSGAN combined with $\alpha = 0.9$. Also, it shows that it did learn to model the activities Walking and Going down stairs better than the activity Running. Experiments conducted on the second generation task were achieved using LSGAN loss function.

### 5.4 Evaluation using Dynamic Time Warping (DTW)

DTW (Sakoe & Chiba, 1978; Vintsyuk, 1968) is a similarity measure between two temporal time series or sequences. For each context, we have the real signal associated with it. We use DTW to compute the similarity between the real and generated signals for each context on the test set. We can compare the trained models by averaging the similarity values obtained on the test dataset. The choice of the DTW is mainly due to the dataset we are working on. We want to detect similarities between the real and generated z-axis series despite the pace of the person's movement. For example, a person can walk faster or slower. The DTW allows this, unlike a point-to-point computing distance like the Euclidean distance.

Table 1: MTS-FID scores of MTS-CGAN trained using different loss functions

| Loss function | MTS-FID | | | |
| --- | --- | --- | --- | --- |
| | All activities | Walking | GDS | Running |
| Standard | 0.23 | 0.34 | 0.14 | 0.42 |
| LSGAN | **0.07** | **0.23** | **0.12** | **0.1** |
| WGAN-GP | 2.81 | 2.58 | 1.82 | 3.89 |

Table 2: MTS-FID scores of MTS-CGAN trained using different alpha scores

| Alpha | MTS-FID | | | |
| --- | --- | --- | --- | --- |
| | All activities | Walking | GDS | Running |
| 0.9 | **0.07** | **0.23** | **0.12** | 0.42 |
| 0.75 | 0.26 | 0.19 | 0.12 | 0.79 |
| .25 | .43 | .5 | 0.53 | 0.99 |
| 0.1 | 0.44 | 0.56 | 0.74 | **.28** |

5.5 DATA AUGMENTATION USING MTS-CGAN

We built a multivariate time series classifier. It classifies the MTS according to their respective ADL (walking, running, going downstairs). We split the dataset into training, validation, and testing sets. We conduct two separate pieces of training. First, we train our model on the original train and validation sets and test on the original set. We balance the training set in the second one by performing a data augmentation technique. We used our MTS-CGAN model to generate the synthetic data. After training, we test the augmented dataset on the original test set. Results showed that enhancing the dataset with our model's synthetic data improved the classifier's performance, especially its accuracy for the class that lacked samples; it went from 77% to 91% in accuracy, as shown in table 4. Note that this method has already been used to evaluate a conditional GAN on multivariate time series (Chen et al., 2021).

Table 3: Training set size before and after balancing the classes using MTS-CGAN

| Class | Walking | GDS | Running |
| --- | --- | --- | --- |
| before | 1132 | 1164 | 782 |
| after | 1132 | 1164 | 1164 |

6 CONCLUSION

In this paper, we presented a novel approach to model multivariate time-series data for conditional generation tasks, which is solely based on Transformer architecture, thus dispensing recurrence entirely. The training process is efficient, and the model simultaneously learned the observed multivariate time series data distribution for each condition. Self-attention mechanisms proved their ability to learn the conditional aspect of the generation while learning the complex dependencies between the multivariate time series. Moreover, we conducted multiple rigorous evaluations in order to validate MTS-CGAN. The qualitative and quantitative evaluation results showed the high quality of the synthetic data and the effectiveness of MTS-CGAN in generating realistic multivariate time series according to the specified condition. This emphasizes the success of attention in conditional modeling of multivariate time series and its potential to model more complex conditions, a direction we plan to explore in the future.

Table 4: Classifier performance before and after data augmentation using MTS-CGAN

|  | Before Data Augmentation | | | | After Data Augmentation | | | |
|---|---|---|---|---|---|---|---|---|
|  | Precision | Recall | F1-score | Acc. | Precision | Recall | F1-score | Acc. |
| Going Down Stairs | 0.77 | 0.77 | 0.77 | 88% | 0.91 | 0.8 | 0.92 | 93% |
| Walking | 0.85 | 0.96 | 0.9 | | 0.87 | 0.99 | 0.92 | |
| Running | 1 | 0.86 | 0.94 | | 1 | 0.96 | 0.98 | |